\documentclass[lettersize,journal]{IEEEtran}
\usepackage{amsmath,amsfonts}
\usepackage{algorithmic}
\usepackage{algorithm}
\usepackage{array}
\usepackage[caption=false,font=normalsize,labelfont=sf,textfont=sf]{subfig}
\usepackage{textcomp}
\usepackage{stfloats}
\usepackage{url}
\usepackage{verbatim}
\usepackage{graphicx}
\usepackage{cite}
\usepackage{verbatim}
\usepackage{multirow}
\usepackage{hyperref}
\begin{document}

\title{GraspGraphNet: Graph-Structured Multi-Embodiment Dexterous Grasp Generation}


\author{Yeonseo Lee$^{1}$,~Taeyeop Lee$^{1}$,~Hyosup Shin$^{1}$,~Guebin Hwang$^{1}$,~and~Sungho Jo$^{1\dagger}$%
\thanks{$^{1}$Korea Advanced Institute of Science and Technology (KAIST), Daejeon, South Korea.}%
\thanks{$^{\dagger}$Corresponding author.}%
}



\maketitle
\begin{abstract}
Dexterous grasp generation across robot hands is challenging because hands differ in kinematic topology, actuation dimensions, and native command spaces.
We introduce GraspGraphNet, a topology-aware grasp generation framework that represents each hand as a URDF-derived kinematic graph and directly generates executable palm poses and joint configurations.
GraspGraphNet combines hierarchical object surface encoding, differentiable forward kinematics, and dynamic world-edge message passing to model evolving robot-object interactions.
It applies conditional flow matching directly in executable palm-pose and joint-state space, avoiding post-processing optimization, inverse kinematics, and retargeting.
Using a shared model trained on Barrett Hand, Allegro Hand, and Shadow Hand, GraspGraphNet achieves an average success rate of 83.48\% with 40\,ms inference time per grasp on a 40-object benchmark.
Without retraining, the same model achieves 72.70\% success on controlled finger-removal variants, demonstrating robustness to hand-topology variations. 
These results suggest that graph-structured hand representations can effectively support dexterous grasp generation across robot hands with different kinematic structures.
Project: \url{https://lysees.github.io/graspgraphnet-page}
\end{abstract}


\section{Introduction}
\IEEEPARstart{D}{exterous} grasping is a fundamental capability for robotic manipulation, requiring a multi-fingered robot hand to coordinate high-dimensional joint motion while establishing physically meaningful contact with object surfaces. 
Recent learning-based methods~\cite{song2025overview, wang2022dexgraspnet, li2023gendexgrasp, zhang2024dexgraspnet, xu2024manifoundation} have made substantial progress in generating stable and diverse grasps for dexterous hands and complex object geometries. 
Despite this progress, scaling dexterous grasp generation across diverse robot hands remains challenging.
Dexterous grasp generation must consider object geometry as well as hand-specific kinematics, link geometry, joint limits, and actuation spaces.

Dexterous robot hands have their own physical characteristics, varying in the number of fingers, link geometry, joint arrangement, and kinematic tree. 
As a result, dexterous robot hands do not share a common configuration space. 
Both the hand representation and the joint command output are hand-specific, so a representation designed for one embodiment is generally not directly applicable to another. 
This makes it difficult to design a single grasping model that supports multiple dexterous hands. 
These challenges highlight the need for a unified model that can operate across robot-hand embodiments with distinct kinematic structures and actuation spaces.

To address this challenge, prior learning-based grasp generation methods~\cite{brahmbhatt2019contactgrasp,li2023gendexgrasp,wei2024dro} have explored transferable representations that reduce dependence on a specific robot configuration space.
Some methods introduce transferable grasp cues, such as object contacts~\cite{brahmbhatt2019contactgrasp, lakshmipathy2022contact}, contact maps~\cite{li2023gendexgrasp, fang2025anydexgrasp}, or contact-centric representations~\cite{xu2024manifoundation, wu2025cedex, khargonkar2025robotfingerprint}.
These representations are transferred to different hands through downstream optimization, retargeting, or hand-specific decision modules.
Other methods instead explicitly model robot-object interaction. DRO-Grasp~\cite{wei2024dro} represents dense robot-object distances, while TRO-Grasp~\cite{fei2025trograspefficientgraph} generates robot-object spatial transformations with graph diffusion.
While such contact and interaction-level representations improve transfer across hands, they are not directly executable on a robot hand, and converting them into executable palm poses and joint commands commonly requires post-processing optimization, inverse kinematics, or retargeting. This conversion increases inference cost and makes the process sensitive to hand kinematics, joint limits, and initialization.

\begin{figure}
\centering
\includegraphics[width=\columnwidth]{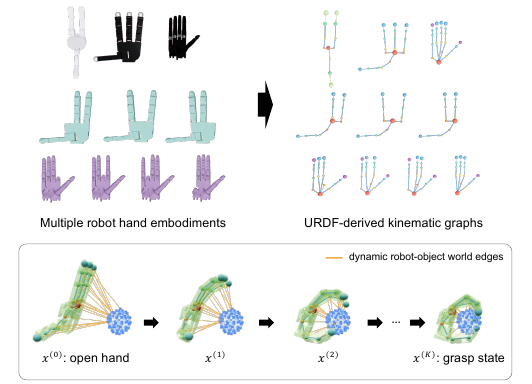}
\caption{
GraspGraphNet represents different dexterous hands as URDF-derived kinematic graphs and generates an executable grasp by evolving an open-hand state $x^{(0)}$ to a final grasp state $x^{(K)}$ with dynamic robot-object world edges.}
\label{fig:figure1}
\end{figure} 

These limitations motivate \textbf{GraspGraphNet}, a topology-aware dexterous grasp generator based on robot-object graph representations.
Rather than predicting an intermediate representation, GraspGraphNet directly generates the palm pose and actuated joint configuration in the configuration space of the input hand.

A robot hand described by a URDF naturally forms an articulated kinematic tree~\cite{tola2024understanding}, in which joints connect physical links.
We represent this URDF structure as a link-joint graph, preserving the articulated structure of the hand so that information can propagate along kinematic connections rather than being represented as an unordered or fixed-length vector.
Graph-based computation enables a single parameter-sharing model to process hand embodiments with different graph sizes, connectivities, and joint dimensions~\cite{battaglia2018relational, wang2018nervenet, patel2025get}, allowing the model to be applied directly to controlled topology variants without retraining.
Building on this graph representation, GraspGraphNet directly outputs the palm pose and joint state of the input hand rather than a hand-agnostic representation that requires a separate reconstruction step.
This keeps grasp generation aligned with each hand's native kinematic structure and reduces inference latency.

Since robot-object proximity changes continuously during grasp formation, the relevant interaction neighborhoods should evolve with the current hand state.
GraspGraphNet therefore constructs dynamic world-frame edges between articulated hand links and nearby object surface points, allowing robot-object interaction features to be updated throughout generation.
We formulate grasp synthesis as conditional flow matching in executable palm-pose and joint-state space.
Starting from an open-hand initialization, the model learns a velocity field that transports the hand toward a final grasp configuration.
Compared with iterative denoising formulations~\cite{ho2020denoising}, flow matching enables efficient trajectory integration with a small number of inference steps, making it well suited for direct executable-state generation~\cite{lipman2022flow}.

We compare GraspGraphNet with prior methods that support multiple robot hands on Barrett, Allegro, and Shadow hands, which span three to five fingers with different kinematic structures.
While prior evaluations~\cite{li2023gendexgrasp, wei2024dro, fei2025trograspefficientgraph} commonly report results on a small set of validation objects, we additionally include 30 objects from Google Scanned Objects~\cite{downs2022google}, resulting in a 40-object evaluation covering a broader range of object geometries.
We further evaluate GraspGraphNet alongside these methods on controlled finger-removal variants that alter the hand graph, link set, and actuated joint dimension.
This evaluation measures robustness to structural variations without retraining.
Experimental results demonstrate that GraspGraphNet achieves strong grasp success rates, efficient inference, and favorable robustness under both expanded object evaluation and finger-removal settings.

The main contributions of this work are summarized as follows:
\begin{itemize}
    \item We propose \textbf{GraspGraphNet}, a topology-aware dexterous grasp-generation framework that operates on URDF-derived hand graphs. Our method supports a unified model across multiple hands and direct application to controlled topology variants without retraining.
    
    \item We introduce a \textbf{dynamic world-edge message passing} mechanism that continuously updates robot-object interaction neighborhoods according to the current articulated hand configuration, enabling the model to capture evolving local contact geometry throughout grasp generation.
    
    \item We formulate dexterous grasp synthesis as \textbf{conditional flow matching in executable palm-pose and joint-state space}, allowing direct generation of grasp commands without embodiment-specific post-processing.
    
\end{itemize}


\section{Related Work}




\subsection{Dexterous Grasp Generation}
Dexterous grasp generation aims to synthesize stable grasp poses and joint configurations for multi-fingered robot hands. 
Early optimization-based methods generate grasps by maximizing force-closure\cite{liu2021synthesizing} or contact-quality objectives\cite{miller2004graspit}, but often require expensive test-time optimization and accurate object geometry. 
To address these limitations, recent learning-based approaches improve efficiency by learning grasp distributions from large-scale datasets\cite{wang2022dexgraspnet, zhang2024dexgraspnet, li2023gendexgrasp}.

Existing learning-based methods can be broadly categorized into robot-centric and object-centric approaches. 
Robot-centric methods directly predict executable grasp states such as wrist poses or joint values, enabling efficient execution but typically restricting the model to a specific hand embodiment and output space\cite{xu2023unidexgrasp, Unidexgrasp++, ye2025dex1b, liu2023dexrepnet}. 
In contrast, object-centric approaches represent grasps through contact points, contact maps, or affordance-related cues, allowing better transfer across objects and robotic hands\cite{shao2020unigrasp, attarian2023geometry, khargonkar2025robotfingerprint, xu2024manifoundation}. 
However, these representations are not directly executable and generally require inverse kinematics, contact fitting, or optimization to recover feasible grasp configurations\cite{shao2020unigrasp, attarian2023geometry, li2023gendexgrasp, khargonkar2025robotfingerprint}. 
This limitation motivates grasp representations that remain transferable across
different robot hands while enabling more direct grasp execution.

\subsection{Dexterous Grasping Across Diverse Robot Hands}
Dexterous grasping across diverse robot hands requires representations that can handle differences in kinematic structure, morphology, and output space while preserving robot-specific grasp feasibility.
A major challenge is to make this representation transferable across embodiments while preserving robot-specific grasp feasibility.

Early approaches primarily relied on contact or geometry-based grasp representations. 
UniGrasp\cite{shao2020unigrasp} and GeoMatch\cite{attarian2023geometry} predict contact points or keypoints conditioned on object and hand geometries, 
while GenDexGrasp\cite{li2023gendexgrasp} and CEDex\cite{wu2025cedex} introduce a hand-agnostic contact-map representation for grasp transfer.
RobotFingerPrint\cite{khargonkar2025robotfingerprint} instead proposes a unified coordinate space that establishes dense correspondences between gripper surfaces and object surfaces. 
These methods improve transferability by avoiding direct prediction in embodiment-specific joint spaces. 
Nevertheless, the predicted contacts, keypoints, or coordinate correspondences must still be converted into executable grasp states through inverse kinematics, optimization, or other hand-specific procedures.

Recent work has shifted toward interaction-centric representations that explicitly model robot-object relationships. 
DRO-Grasp\cite{wei2024dro} represents grasps through robot-object distance relationships, 
while TRO-Grasp\cite{fei2025trograspefficientgraph} further introduces graph-based transformation representations between object patches and hand links. 
These methods demonstrate that robot-object interactions provide effective transferable representations across embodiments, but they still require a separate reconstruction or optimization stage to obtain executable robot commands, such as palm poses and joint configurations.
Our work follows this interaction-centric direction but directly generates palm poses and joint configurations on a URDF-derived hand graph.
Furthermore, robot-object interaction edges are dynamically updated during generation, enabling the interaction structure to evolve with the articulated hand state.

\section{Proposed Method}
\label{sec:method}
\begin{figure*}
\centering
\includegraphics[width=2\columnwidth]{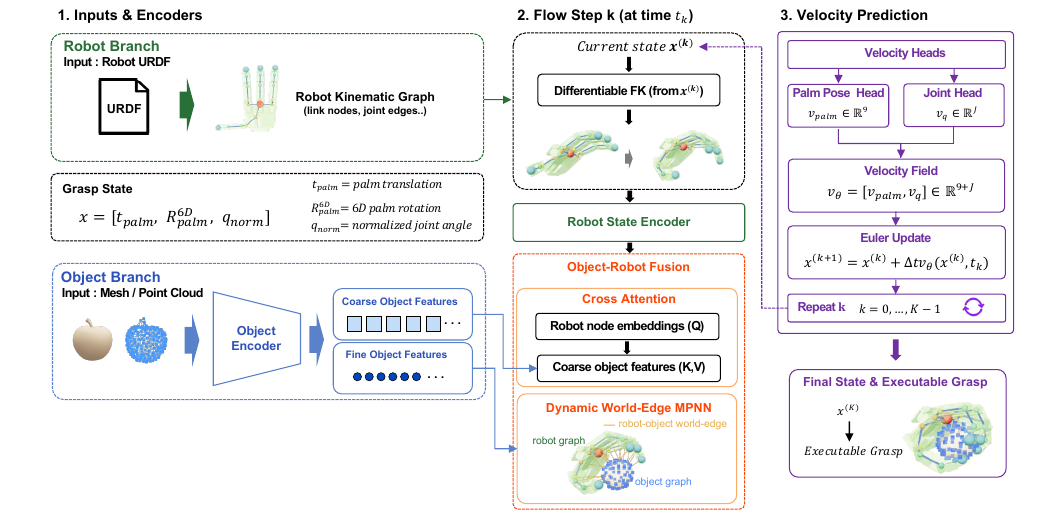}
\caption{Overview of GraspGraphNet.
Given an object point cloud and a robot kinematic graph constructed from its URDF, GraspGraphNet generates an executable grasp by integrating a learned velocity field over the palm pose and joint state.
At each flow step, differentiable forward kinematics computes the current world-space hand-link poses, and a kinematic-tree MPNN encodes the articulated robot state.
The resulting robot node embeddings are fused with coarse and fine object features through cross-attention and dynamic world-edge message passing,
respectively.
Palm and joint velocity heads then predict the velocity field used to update the grasp state.}
\label{fig:figure2}
\vspace{-10pt}
\end{figure*} 
\subsection{Overview and Grasp-State Formulation}
We propose GraspGraphNet, a topology-aware dexterous grasp generator that operates directly on object and robot graph representations.
Given a target object point cloud and a robot URDF, 
the model generates an executable grasp consisting of the palm pose and actuated joint configuration of the input hand.
Unlike methods that first predict an interaction representation and then convert it into robot commands, 
GraspGraphNet learns a conditional velocity field directly over the executable grasp-state space.

For a robot hand with $J$ actuated joints, we represent a grasp state as
\begin{equation}
    x =
    \left[
        t_{\mathrm{palm}},
        R_{\mathrm{palm}}^{6D},
        q_{\mathrm{norm}}
    \right]
    \in \mathbb{R}^{9+J},
    \label{eq:grasp_state}
\end{equation}
where $t_{\mathrm{palm}}\in\mathbb{R}^{3}$ is the palm translation in the object-centered frame, $R_{\mathrm{palm}}^{6D}\in\mathbb{R}^{6}$ is the continuous 6D palm-orientation representation, 
and $q_{\mathrm{norm}}\in\mathbb{R}^{J}$ contains the normalized actuated joint angles.
To provide consistent orientation semantics across embodiments, the palm orientation is represented in a robot independent canonical palm frame.
Each joint is normalized independently using the limits specified in the robot URDF.
Because $J$ is determined by the input robot graph, the grasp-state dimension can vary across hand embodiments.

As illustrated in Fig.~\ref{fig:figure2}, the object and robot representations condition a state-dependent vector field.
At each flow step, the current state is articulated through differentiable forward kinematics (FK), encoded over the robot kinematic tree, and fused with the object representation.
Palm and joint velocity heads then predict an update to the current grasp state.

\subsection{Object and Robot Graph Representations}
GraspGraphNet uses complementary graph representations for the target object and the robot hand.
The object representation captures geometry at multiple spatial resolutions,
whereas the robot representation preserves the URDF-defined link-joint topology and link geometry of the input hand.
Object points and state-conditioned robot poses are expressed in an
object-centered world frame, while link geometry is initially represented in each link's local frame.

\subsubsection{Object representation}
Given an object mesh or point cloud, we sample $1{,}024$ surface points, each represented by its position, surface normal, and local curvature.
We construct a four-level hierarchy with $1{,}024$, $256$, $64$, and $32$ points, where farthest-point sampling produces coarser levels and local features are aggregated from the preceding level~\cite{qi2017pointnet++}.
A $k$-nearest-neighbor graph is built within each level for hierarchical point-graph message passing.
We use the $64$-point embeddings as fine object features $O_f$ for contact-relevant geometry and the $32$-point embeddings as coarse object features $O_c$ for global shape context.
Mean pooling the coarse features yields a global object descriptor for the grasp-velocity heads.

\subsubsection{Robot representation}
\begin{figure}
\centering
\includegraphics[width=\columnwidth]{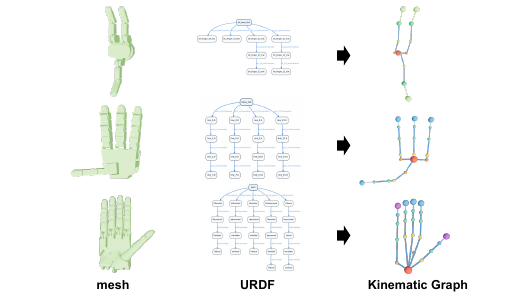}
\caption{
URDF-derived robot hand representations.
Physical links form graph nodes, and joints form directed parent-to-child
edges.
}
\label{fig:figure3}
\vspace{-10pt}
\end{figure} 
As illustrated in Fig.~\ref{fig:figure3}, we construct a robot kinematic graph $\mathcal{G}_R=(\mathcal{V}_R,\mathcal{E}_R)$ directly from the input URDF.
Each node $i\in\mathcal{V}_R$ corresponds to a physical robot link, and each directed edge $(i,j)\in\mathcal{E}_R$ corresponds to a joint connecting a parent link $i$ to a child link $j$.
Virtual and robot-specific unused links are excluded.
This representation naturally changes in graph size and connectivity across robot embodiments.
Each kinematic edge stores its joint type, joint axis, joint limits, and fixed parent-to-child transformation.
To represent link geometry, we sample points from each visual mesh and encode them using a shared PointNet to obtain a per-link geometry embedding.
These geometry embeddings are included in the state-conditioned robot node features.

\subsection{State-Conditioned Robot Encoding}
At flow time $t$, the current state $x_t$ specifies a palm pose and normalized joint configuration.
We denormalize the joint values using the URDF limits and apply differentiable forward kinematics over $\mathcal{G}_R$ to compute the world-space transformation $T_i^t$ of each robot link.
The link position $p_i^t$ and orientation $R_i^t$ are extracted from $T_i^t$.
For each robot link, we construct a state-conditioned node feature by concatenating its link-geometry embedding, current FK-derived position and 6D orientation, kinematic-tree depth embedding, fingertip indicator, and the normalized state of its associated joint.
For links not associated with an actuated joint, the joint-state input is set to zero.
The concatenated feature is then projected to the robot-node embedding space using an MLP.

The projected node features are processed by a three-layer edge-conditioned Message Passing Neural Network (MPNN) over the URDF-derived kinematic edges.
Messages are conditioned on the joint attributes stored on each kinematic edge.
Consequently, the encoder captures the current articulated hand state while preserving the topology and link geometry of the input embodiment.

\subsection{Object-Robot Fusion}
\subsubsection{Coarse Cross-Attention}
The robot node embeddings serve as queries, while the coarse object features serve as keys and values.
This operation conditions the articulated robot state on global object-shape context.

\subsubsection{Dynamic World-Edge Message Passing}
For each robot link, we then identify the $K_{\mathrm{W}}$ nearest fine object points and construct dynamic edges from these points to the robot node:
\begin{equation}
    \mathcal{E}_W^t
    =
    \operatorname{kNN}
    \left(
        \{p_j^{\mathrm{obj}}\},
        \{p_i^t\}
    \right).
\end{equation}
Here, $p_j^{\mathrm{obj}}$ denotes a fine-level object point and $p_i^t$ denotes the current world-space position of robot link $i$.
This produces $K_{\mathrm{W}}|\mathcal{V}_R|$ directed edges and scales linearly with the number of robot links.
Each message combines the fine object feature, current robot embedding, 
relative displacement, and distance. 
The messages from neighboring object points are aggregated at each robot node through residual update layers.
Rebuilding $\mathcal{E}_W^t$ at every flow step allows the local interaction graph to evolve with the generated grasp trajectory.

\subsection{Velocity Field and Flow-Matching Objective}
\subsubsection{Velocity prediction}
The object-aware robot embeddings are mapped to palm and joint velocities.
The palm pose and joint heads shown in Fig.~\ref{fig:figure2} predict the palm-state velocity $v_{\mathrm{palm}}$ and per-joint velocity $v_q$, respectively.
We designate the link node with the highest out-degree in the kinematic graph as the palm-hub node, which generally corresponds to the physical palm from which the finger chains branch.
The palm velocity head reads the embedding of the designated palm-hub node, whereas the joint velocity head is shared across all actuated joints and is applied to the child-link embedding associated with each joint.
Both heads are additionally conditioned on the global object descriptor and a sinusoidal time embedding.
The scalar predictions are assembled according to the actuated-joint indices of the input graph to form $v_q\in\mathbb{R}^{J}$.
The final grasp-state velocity is
\begin{equation}
    v_\theta(x_t,t)
    =
    [v_{\mathrm{palm}};v_q]
    \in\mathbb{R}^{9+J}.
    \label{eq:velocity_field}
\end{equation}
The shared per-joint head allows the same network to operate on hands with different actuation dimensions.
We use $x_t$ to denote a state at continuous flow time $t$ during training,
and $x^{(k)}$ to denote its numerical approximation at the $k$-th Euler step during inference, where $t_k=k/K$.

\subsubsection{Initial and target states}
For each training example, the target state $x_1$ is obtained from a
ground-truth grasp annotation.
The initial state $x_0$ is an open-hand pose sampled around the object, with
all actuated joints initialized to the open-hand configuration.
During training, we sample the initial palm approach direction within a cone
centered at this target direction and sample the in-plane palm rotation
uniformly over $[0,2\pi)$.

When multiple ground-truth grasps are available for the same robot-object
pair, we choose the target grasp that is closest to the sampled initial state.
The matching cost combines palm-translation and palm-rotation differences.
By pairing each sampled initial state with a nearby target grasp, the model is
trained to move the hand toward a compatible grasp configuration from diverse
starting poses.
At inference time, no target grasp is available, so we sample the initial palm
approach direction uniformly on the sphere while keeping the same open-hand
joint initialization.

\subsubsection{Flow-matching objective}
We sample $t\sim\mathcal{U}(0,1)$ and linearly interpolate between the initial
and target states, yielding the target conditional velocity
\begin{equation}
    x_t=(1-t)x_0+t x_1,
    \qquad
    v^*=x_1-x_0.
\end{equation}
This standard linear conditional path is defined in the continuous grasp-state representation, using a continuous 6D orientation representation and joint-limit-normalized joint values.
GraspGraphNet predicts $v_\theta(x_t,t)$, and we optimize the velocity-matching objective
\begin{equation}
    \mathcal{L}
    =
    \operatorname{MSE}
    \left(
        v_{\mathrm{palm}},
        v_{\mathrm{palm}}^*
    \right)
    +
    \lambda_q
    \operatorname{MSE}
    \left(
        v_q,
        v_q^*
    \right).
\end{equation}
Here, $v_{\mathrm{palm}}^*$ and $v_q^*$ are the palm and joint components of $v^*$, and $\lambda_q$ balances their supervision.

\subsection{Grasp Generation}
\begin{figure}
\centering
\includegraphics[width=0.9\columnwidth]{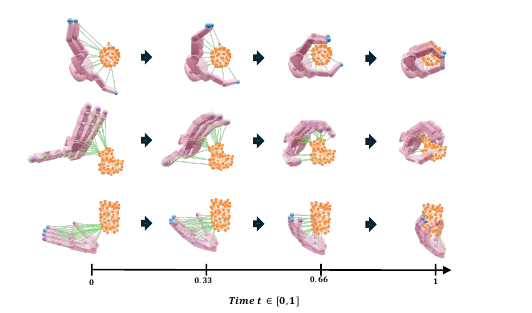}
\caption{State-dependent grasp generation. The grasp state is iteratively refined from an open-hand initialization with state-dependent forward kinematics and dynamic world-edge updates.}
\label{fig:figure_world_edge}
\vspace{-10pt}
\end{figure} 

At inference time, we sample an initial open-hand state $x^{(0)}$ and integrate the learned velocity field using $K$ Euler steps:
\begin{equation}
\begin{aligned}
    x^{(k+1)}
    &=
    x^{(k)}
    +
    \Delta t\,
    v_\theta
    \left(
        x^{(k)},t_k
    \right),
    \\
    t_k &= \frac{k}{K},
    \qquad
    \Delta t=\frac{1}{K}.
\end{aligned}
\label{eq:euler_inference}
\end{equation}
At every step, the updated grasp state changes the articulated hand geometry and therefore the dynamic object-robot world edges.
Unless otherwise specified, we use $K=3$, which the integration-step ablation shows to provide a favorable trade-off between grasp success and inference time.

The final state $x^{(K)}$ specifies a palm pose and normalized actuated-joint configuration.
For execution, the canonical palm orientation is converted to the robot-native frame, the fixed palm-to-wrist offset defined by the robot model is applied,
and the joint values are denormalized using their URDF limits.
This produces an executable wrist pose and physical joint commands for the input robot hand.

\section{Experiments}
We evaluate GraspGraphNet to examine whether a single model can generate
stable executable grasps for robot hands with different kinematic structures.
The simulation experiments measure grasp success and inference time across Barrett, Allegro, and Shadow hands. 
The finger-removal experiments evaluate robustness to hand-topology changes. 
The real-world experiments with a Leap Hand assess whether the model can reliably generate grasps from sensor-captured object observations in a physical robotic setup.
We also perform ablation studies to analyze the contributions of object-robot fusion, dynamic world-edge updates, integration steps, and the flow-matching objective.

\subsection{Experimental Setup}
\paragraph{\textbf{Training Dataset}}
GraspGraphNet is trained on filtered CMapDataset~\cite{li2023gendexgrasp}, following the multi-hand evaluation setting of prior dexterous grasping
works~\cite{wei2024dro, fei2025trograspefficientgraph}. 
The dataset contains object point clouds and grasp annotations for multiple dexterous robot hands. 
For fair comparison, all methods are trained under the same three-hand setting, using Barrett Hand, Allegro Hand, and Shadow Hand data from the same CMapDataset training split.

\begin{figure}
\centering
\includegraphics[width=0.9\columnwidth]{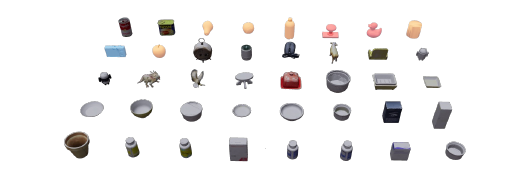}
\caption{Evaluation object set used in simulation. The set contains 40 unseen objects, including 10 held-out CMapDataset objects and 30 objects from Google Scanned Objects.}
\label{fig:Figure_object}
\vspace{-10pt}
\end{figure}

\paragraph{\textbf{Simulation Evaluation Protocol}}
Generated grasps are evaluated in Isaac Gym using the same simulation-based protocol as DRO-Grasp~\cite{wei2024dro} and TRO-Grasp \cite{fei2025trograspefficientgraph}. 
The evaluation object set consists of 40 unseen objects, including 10 objects held out from CMapDataset and 30 objects selected from Google Scanned Objects~\cite{downs2022google}. 
Fig.~\ref{fig:Figure_object} shows the full set of evaluation objects.
For each robot hand, 100 grasp trials are generated for each object, resulting in 4000 evaluation trials per hand.
A grasp is considered successful if the object displacement remains below 2 cm under six directional perturbations in Isaac Gym. 
The success rate is reported as the percentage of successful trials over all generated grasps. 
Inference time is measured as the computational time required to generate one executable grasp command, excluding physics simulation.
For a fair comparison, all methods are evaluated using the same object set, simulation environment, evaluation protocol, and hardware.

\paragraph{\textbf{Implementation Details}}
GraspGraphNet uses a hidden dimension of 128 and approximately 1.58 million parameters.
The object graph contains four levels with 1024, 256, 64, and 32 points, while each robot link is represented by 64 sampled points.
We use three robot-graph layers, two world-edge message-passing layers, and $k=8$ nearest object points per robot link.
The model is trained for 200 epochs using AdamW with an initial learning rate of $3\times10^{-4}$, cosine decay, and $\lambda_q=1$.
For initial-state sampling and target-grasp selection, we use a cone angle of $30^\circ$ and a rotation-cost weight of 0.5, respectively.
Unless otherwise specified, inference uses $K=3$ Euler steps.
All experiments are conducted on a single NVIDIA RTX 4090 GPU.

\subsection{Simulation Results}
\label{sec:overall_performance}
\begin{table*}[t]
\centering
\caption{Comparison of success rate and inference time across methods.}
\label{tab:comparison}
\resizebox{\textwidth}{!}{
\begin{tabular}{l|cccc|cccc}
\hline
\multirow{2}{*}{Method} &
\multicolumn{4}{c|}{Success Rate (\%) $\uparrow$} &
\multicolumn{4}{c}{Inference Time (ms) $\downarrow$} \\
\cline{2-9}
& Barrett & Allegro & ShadowHand & Avg.
& Barrett & Allegro & ShadowHand & Avg. \\
\hline
GenDexGrasp~\cite{li2023gendexgrasp}
& 38.57(1543/4000) & 19.72(789/4000) & 28.87(1155/4000) & 29.05
& 11070 & 19750 & 15840 & 15550 \\

DRO-Grasp~\cite{wei2024dro}
& 85.65(3426/4000) & 81.07(3243/4000) & 66.07(2643/4000) & 77.60
& 583 & 421 & 703 & 569 \\

TRO-Grasp~\cite{fei2025trograspefficientgraph}
& 82.12(3285/4000) & 86.57(3463/4000) & 74.30(2972/4000) & 81.00
& 264 & 265 & 267 & 265 \\

\textbf{GraspGraphNet (Ours)}
& \textbf{86.05(3442/4000)} & \textbf{87.75(3510/4000)} & \textbf{76.65(3066/4000)} & \textbf{83.48}
& \textbf{24} & \textbf{33} & \textbf{63} & \textbf{40} \\
\hline
\end{tabular}
}
\vspace{-10pt}
\end{table*}

This experiment measures the success rate and inference time of a single model across robot hands with different kinematic structures.
Table~\ref{tab:comparison} reports grasp success rate and inference time on Barrett Hand, Allegro Hand, and Shadow Hand.
GraspGraphNet achieves the highest average success rate among all compared methods, reaching 83.48\%.
Compared with recent multi-hand grasp generation baselines, this improves the average success rate by 5.88\% over DRO-Grasp and 2.48\% over TRO-Grasp.
GenDexGrasp performs lower in this evaluation, suggesting that recovering executable robot-specific grasps from hand-agnostic contact maps remains challenging across diverse hand embodiments.
Across individual hands, GraspGraphNet remains competitive or superior on all three embodiments, achieving the best success rate on Barrett Hand, Allegro Hand, and Shadow Hand.

GraspGraphNet also provides a substantial improvement in inference efficiency.
GenDexGrasp has the highest inference cost because each grasp requires contact-map-guided particle optimization.
DRO-Grasp predicts an interaction representation rather than executable robot states, requiring additional reconstruction and optimization stages to obtain robot-specific grasp configurations.
TRO-Grasp generates intermediate robot-object transformation graphs through diffusion and subsequently solves inverse kinematics to recover executable joint configurations.
In contrast, GraspGraphNet directly integrates a learned velocity field in executable palm-pose and joint-state space, producing robot-specific grasp commands.
As a result, GraspGraphNet requires only 40 ms per grasp on average, compared with 15550 ms for GenDexGrasp, 569 ms for DRO-Grasp, and 265 ms for TRO-Grasp.
This corresponds to approximately $14\times$ faster inference than DRO-Grasp and $6\times$ faster inference than TRO-Grasp under the same evaluation setting.
These results indicate that GraspGraphNet achieves both strong grasp success and high inference efficiency across multiple dexterous robot hand embodiments.

\subsection{Generalization to Topology-Modified Hands}
\begin{figure}
\centering
\includegraphics[width=0.95\columnwidth]{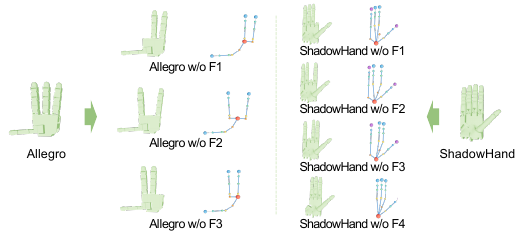}
\caption{Finger-ablated robot hands used for topology generalization. Each variant is represented by its URDF-derived kinematic graph.}
\label{fig:removed_finger}
\vspace{-10pt}
\end{figure}

To evaluate generalization to hand-topology modifications, we construct finger-ablated hands by removing one finger from each original hand.
Three finger-ablated variants are constructed for Allegro Hand and four variants for Shadow Hand, as illustrated in Fig.~\ref{fig:removed_finger}.
All methods are evaluated using the same models as in Sec.~\ref{sec:overall_performance}, without any retraining or adaptation. Removing a finger changes the URDF kinematic tree, the number of links, and the number of actuated joints, making this setting a direct evaluation of robustness to topology modifications of the hand embodiment.

\begin{figure}
\centering
\includegraphics[width=\columnwidth]{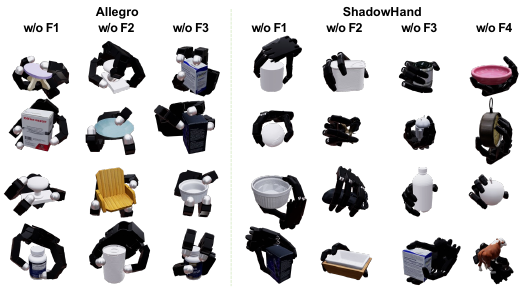}
\caption{Qualitative grasp results on topology-modified robot hands.}
\label{fig:finger_grasp_results}
\vspace{-10pt}
\end{figure} 
\begin{table*}[t]
\centering
\caption{Comparison of success rates under finger-removal settings across methods.}
\label{tab:ablation_finger_removal}

\small
\setlength{\tabcolsep}{5pt}

\begin{tabular}{l|ccc|cccc|c}
\hline
\multirow[c]{2}{*}{Method} &
\multicolumn{3}{c|}{Allegro} &
\multicolumn{4}{c|}{ShadowHand} &
\multirow{2}{*}{Overall} \\
\cline{2-8}

& w/o F1 & w/o F2 & w/o F3
& w/o F1 & w/o F2 & w/o F3 & w/o F4
& Avg. \\
\hline
GenDexGrasp\cite{li2023gendexgrasp} 
& 20.75 & 24.02 &  17.00 & 6.95 & 8.30 & 7.55 & 6.37 & 12.99 \\
DRO-Grasp\cite{wei2024dro} 
& 63.00 & 64.90 &  \textbf{58.15}& 43.22& 47.40& 62.75& \textbf{58.05}& 56.78 \\
TRO-Grasp\cite{fei2025trograspefficientgraph} 
& 21.47 &  18.80& 9.10& 5.92& 9.60& 30.25& 11.95& 15.29 \\
\textbf{GraspGraphNet (Ours)}
& \textbf{87.40}& \textbf{84.07}& 57.45& \textbf{73.72}& \textbf{75.12}& \textbf{75.80}& 55.35& \textbf{72.70} \\
\hline
\end{tabular}
\vspace{-10pt}
\end{table*}
Table~\ref{tab:ablation_finger_removal} reports success rates on the finger-ablated hands.
GraspGraphNet achieves the highest overall success rate of 72.70\%, outperforming DRO-Grasp by 15.92\% and substantially outperforming GenDexGrasp and TRO-Grasp.
GenDexGrasp exhibits a substantial performance drop under finger removal. 
Because it relies on predicted contact maps, some learned contacts become infeasible after finger removal.
TRO-Grasp also shows a noticeable decrease in performance. Its graph-based transformation representation is defined over object patches and hand links, making the learned interaction structure more sensitive to changes in the underlying hand topology. In contrast, DRO-Grasp remains relatively robust, suggesting that dense robot-object distance relationships provide a more transferable interaction representation under moderate topology variations.

GraspGraphNet achieves the best performance in five of the seven finger-removal settings and remains competitive in the remaining two, where the gap to DRO-Grasp is less than 3\%. Since robot embodiments are represented as URDF-derived kinematic graphs, changes in the number of links, joints, and graph connectivity can be directly incorporated into the model input without modifying the network architecture. This enables the learned grasp generation process to better accommodate topology variations across robot hands.
Qualitative results in Fig.~\ref{fig:finger_grasp_results} further illustrate that GraspGraphNet remains effective despite changes in hand topology.

\subsection{Real-World Experiments }
\begin{figure}
\centering
\includegraphics[width=0.90\columnwidth]{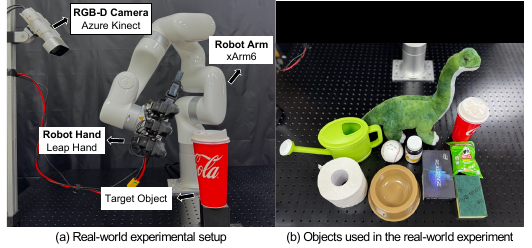}
\caption{Real-world experimental setup and test objects. (a) Experimental setup with the RGB-D camera, xArm6 robot arm, and Leap Hand. (b) Objects used in the real-world experiments.}
\label{fig:real_world}
\end{figure} 
We evaluate GraspGraphNet in real-world experiments, as shown in Fig.~\ref{fig:real_world}. 
The system consists of a UFactory xArm6 robot arm equipped with a Leap Hand~\cite{shaw2023leaphand} and an Azure Kinect RGB-D camera. 
For each trial, a partial point cloud of the target object is obtained from the RGB-D camera and provided as input to GraspGraphNet. 
The generated executable palm pose and joint commands are then directly executed on the robot.
For real-world deployment, we fine-tune GraspGraphNet using the Leap Hand training data released by TRO-Grasp~\cite{fei2025trograspefficientgraph}. 
We evaluate GraspGraphNet on 10 objects with diverse shapes and sizes, shown in Fig.~\ref{fig:real_world}(b). 
For each object, we perform 10 grasp trials, resulting in 100 real-world grasp attempts in total.
A trial is considered successful if the robot grasps the object, lifts it from the table, and maintains the grasp without dropping it. 
GraspGraphNet achieves a success rate of 91\%, demonstrating robust grasp generation from partial real-world point clouds and direct execution on a physical robot.

\subsection{Ablation Study}
\subsubsection{Flow Matching vs. Direct Regression}
\begin{table}[t]
\centering
\caption{Ablation on direct regression and flow matching.}
\label{tab:flow_vs_direct}

\small
\setlength{\tabcolsep}{4pt}

\begin{tabular}{l|cccc}
\hline
\multirow{2}{*}{Method} &
\multicolumn{4}{c}{Success Rate (\%) $\uparrow$} \\
\cline{2-5}

& Barrett & Allegro & ShadowHand & Avg. \\
\hline
Direct Regression      & 62.45 & 61.15 & 62.33 & 61.98 \\
Flow Matching ($K=1$)  & 82.75 & 88.02 & 66.05 & 78.94 \\
\textbf{Flow Matching ($K=3$)} & \textbf{86.05} & \textbf{87.75} & \textbf{76.65} & \textbf{83.48} \\
\hline
\end{tabular}
\vspace{-10pt}
\end{table}
We compare conditional flow matching with a direct-regression variant under the same experimental setting, changing only the generation objective. 
The direct-regression model predicts the final displacement from the initial hand state in a single forward pass, rather than learning a time-conditioned velocity field.

As shown in Table~\ref{tab:flow_vs_direct}, direct regression achieves an average success rate of 61.98\%, substantially lower than flow matching with one Euler step (78.94\%) and three steps (83.48\%). 
This suggests that dexterous grasp generation is difficult to model as a single-step regression problem due to the multimodality of the grasp distribution, since multiple palm poses and joint configurations may correspond to valid grasps for the same robot-object pair.

Direct regression can therefore produce averaged or imprecise grasp configurations. 
In contrast, flow matching models grasp generation as a sequence of state-conditioned updates, allowing the hand pose and joint configuration to be progressively refined toward stable grasps.

\subsubsection{Object-Robot Fusion Modules}
\begin{table}[t]
\centering
\caption{Ablation on object-robot fusion modules.}
\label{tab:ablation_world_edge}

\small
\setlength{\tabcolsep}{4pt}

\begin{tabular}{l|cccc}
\hline
Method
& Barrett
& Allegro
& ShadowHand
& Avg. \\
\hline

w/o Cross-Attention
&  72.90&  48.37&  43.15&  54.80\\

w/o World-Edge MPNN
&  52.30&  50.05&  51.47&  51.27\\

\textbf{Full Model}
&  \textbf{86.05}&  \textbf{87.75}&  \textbf{76.65}&  \textbf{83.48}\\

\hline
\end{tabular}
\vspace{-10pt}
\end{table}
We ablate the two object-robot fusion modules in GraspGraphNet, cross-attention and world-edge message passing, to examine their complementary roles. 
Cross-attention provides global object context, while world-edge message passing captures local contact-relevant geometry in the shared world frame. 
Removing cross-attention significantly reduces the average success rate from 83.48\% to 54.80\%. 
This indicates that global object context is important for conditioning robot node features on the overall shape of the target object. 
Removing the World-Edge MPNN further reduces the average success rate to 51.27\%, demonstrating the importance of modeling local hand-object geometric interactions during grasp generation.
Our full model, which combines cross-attention and dynamic world-edge message passing, achieves the best performance across all three robot hands.

\subsubsection{Effect of dynamic world edges}
\begin{table}[t]
\centering
\caption{Ablation study on dynamic world-edge message passing.}
\label{tab:ablation_dynamic_edge}

\small
\setlength{\tabcolsep}{4pt}

\begin{tabular}{l|cccc}
\hline
\multirow{2}{*}{Method} &
\multicolumn{4}{c}{Success Rate (\%) $\uparrow$} \\
\cline{2-5}

& Barrett & Allegro & ShadowHand & Avg. \\
\hline
Static world edges      & 72.52 & 42.55 & 35.62 & 50.23 \\
\textbf{Dynamic world edges} & \textbf{86.05} & \textbf{87.75} & \textbf{76.65} & \textbf{83.48} \\
\hline
\end{tabular}
\end{table}
We further analyze the world-edge message-passing module used for object-robot fusion. 
In the full model, robot-object world edges are recomputed at every flow step between the current FK-derived robot link positions and nearby object surface points. 
To isolate the effect of this dynamic connectivity, we compare against a static-edge variant in which these world edges are constructed once from the initial open-hand state $x_0$ and reused throughout flow integration. 
All other components, including the hand state, FK-derived link poses, robot node features, and message-passing updates, remain unchanged. 
As shown in Table~\ref{tab:ablation_dynamic_edge}, static world edges reduce the average success rate from 83.48\% to 50.23\%, with especially large drops for Allegro Hand and Shadow Hand. 
This indicates that fixed interaction neighborhoods cannot reliably track the changing local robot-object geometry during grasp formation. 
The results highlight the importance of updating robot-object interaction neighborhoods as the hand configuration evolves throughout grasp generation.

\subsubsection{Integration Steps $K$}

\begin{table}[t]
\centering
\caption{Effect of integration steps $K$ on success rate and inference time.}
\label{tab:ablation_k}

\small
\setlength{\tabcolsep}{4pt}

\begin{tabular}{c|cccc|c}
\hline
\multirow{2}{*}{$K$} &
\multicolumn{4}{c|}{Success Rate (\%) $\uparrow$} &
\multirow{2}{*}{Inference Time (ms) $\downarrow$} \\
\cline{2-5}

& Barrett & Allegro & ShadowHand & Avg. & \\
\hline
1  &  82.75&  88.02&  66.05&  78.94&  9\\
2  &  84.95&  88.25&  74.00&  82.40&  19\\
\textbf{3}  &  86.05&  87.75&  76.65&  83.48&  40\\
5  &  85.97&  88.05&  76.07&  83.36&  47\\
10 &  86.87&  87.70&  76.45&  83.67&  95\\
20 &  87.07&  88.65&  75.90&  83.87&  189\\
\hline
\end{tabular}
\vspace{-10pt}
\end{table}
We evaluate the effect of the number of Euler integration steps $K$ used during inference.
As shown in Table~\ref{tab:ablation_k}, increasing $K$ from 1 to 3 improves the average success rate from 78.94\% to 83.48\%. 
This suggests that multiple flow updates help refine the palm pose and joint configuration as the hand approaches the object. 
However, further increasing $K$ provides only marginal gains. Increasing $K$ from 3 to 20 improves the average success rate by only 0.39\%, while increasing inference time from 40 ms to 189 ms per grasp.
We therefore use $K=3$ in all main experiments as a practical trade-off between grasp quality and inference time.

\section{Conclusion}
In this paper, we presented GraspGraphNet, a graph-based framework for generating executable dexterous grasps across robot hands with different kinematic structures and actuation spaces.
By representing robot hands as URDF-derived kinematic graphs and modeling robot-object interactions, GraspGraphNet enables grasp generation across multiple dexterous hand embodiments with different kinematic structures.
To address the multimodality of dexterous grasp generation, 
we formulate grasp synthesis as a conditional flow matching problem that iteratively refines palm poses and joint configurations in an executable grasp state space. 
Unlike prior approaches that rely on contact-based representations or grasp targets requiring additional optimization, retargeting, or inverse kinematics, GraspGraphNet directly generates executable palm poses and joint commands. 
Experimental results demonstrate strong performance across multiple dexterous hands, robustness to topology variations, and successful deployment on a real-world robotic system.
Future work will investigate zero-shot generalization beyond topology variants to robot hands with substantially different morphologies, and extend the framework to broader dexterous manipulation tasks.

\bibliographystyle{IEEEtran}
\bibliography{IEEEabrv,ref}

\end{document}